\documentclass[letterpaper]{article} 
\usepackage{aaai2026}  
\usepackage{times}  
\usepackage{helvet}  
\usepackage{courier}  
\usepackage[hyphens]{url}  
\usepackage{graphicx} 
\usepackage{natbib}  
\usepackage{caption} 
\usepackage{algorithm}
\usepackage{algorithmic}

\usepackage{amssymb}
\usepackage{algorithm}
\usepackage{algorithmic}
\usepackage{tabularx}
\usepackage{booktabs} 
\usepackage{newfloat}
\usepackage{listings}
\usepackage{multirow}
\usepackage{graphicx}
\usepackage{amsmath}
\usepackage{xcolor}

\usepackage{newfloat}
\usepackage{listings}
\DeclareCaptionStyle{ruled}{labelfont=normalfont,labelsep=colon,strut=off} 
\floatstyle{ruled}
\newfloat{listing}{tb}{lst}{}
\floatname{listing}{Listing}
\title{A Lightweight Plastic-Memory Framework for Graph Few-Shot Class-Incremental Learning}
\author{
    Zihan Mei,
    Zhili Qin,
    Tongze Zhang,
    Hongyuan Liu,
    Junming Shao\thanks{Corresponding author.},
    Qinli Yang
}
\affiliations{
    University of Electronic Science and Technology of China\\
    \{zihanmei, zhangtongze, hongyuanliu, junmshao, qlyang\}@uestc.edu.cn, qinzhili@outlook.com

}
\nocopyright
\usepackage{bibentry}

\begin{document}

\maketitle

\begin{abstract}
Graph Incremental Learning has garnered increasing attention as dynamic graph data continues to emerge across diverse fields.  Conventional approaches primarily address catastrophic forgetting by preserving node-related knowledge through replay or distillation techniques;  however, they often incur high computational costs and inefficiency.  This issue is further exacerbated in real-world scenarios where labeled data for new classes is scarce.  In this paper, we propose a novel lightweight plastic-memory framework specifically designed for few-shot incremental learning on graphs.  The core idea of our framework is the construction of a plastic-memory module that evolves over time, continuously updating and expanding its memory to accommodate new classes while retaining previously learned knowledge.  In contrast to existing techniques, our memory module is both lightweight and effective, featuring an innovative evolving micro-clustering structure that dynamically updates representations of class prototypes, sub-prototypes, and their interaction weights.  Building on this memory module, we introduce a memory-driven meta-learning framework that enhances adaptability to new tasks in its inner loop while maintaining stability for earlier tasks in the outer loop.  Extensive experiments on four benchmark datasets demonstrate the framework's superior performance in balancing stability for old knowledge and adaptability to new knowledge.
\end{abstract}


\section{Introduction}

Graph Incremental Learning has emerged as a critical area of research due to the increasing prevalence of dynamic graph data in real-world applications, such as social networks, recommendation systems, and biological networks \cite{9416834,yuan2024survey}. These domains often involve evolving structures and relationships, requiring models that can adapt to new information without forgetting prior knowledge. A key challenge is the scarcity of labeled data for new classes, making effective training difficult. Few-shot learning, which generalizes from limited labeled examples, becomes crucial in graph-based tasks where extensive annotations are costly or impractical. As a result, integrating few-shot learning with graph incremental learning—Graph Few-Shot Class-Incremental Learning (GFSCIL)—has emerged as an essential research direction. GFSCIL requires models to learn distinct node features with scarce labeled samples while retaining old knowledge and quickly adapting to new information. This scenario involves addressing the stability-plasticity dilemma, balancing the prevention of catastrophic forgetting with the efficient integration of new knowledge.

Despite significant progress in graph incremental learning, existing methods face critical limitations in addressing GFSCIL, necessitating more flexible and lightweight frameworks. Traditional methods mitigate forgetting by explicitly storing old-class samples or fixing feature space topology, which leads to high memory consumption \cite{snell2017prototypical, zhou2022few, zhou2021overcoming}. Additionally, graph construction and updates are highly sensitive to few-shot data distributions, with limited labeled samples making the topology prone to noise \cite{kim2019edge, tian2024survey, zhou2022few}, and traditional knowledge distillation methods exacerbating forgetting due to extreme class imbalance \cite{dong2021few, tao2020few}. Furthermore, most frameworks lack adaptability to dynamic incremental scenarios, relying on fixed network structures or complex multi-stage training strategies that are ill-suited for evolving class streams \cite{kim2019edge, tian2024survey}, while meta-learning-based dynamic networks face challenges in balancing forgetting and computational costs \cite{chi2022metafscil}. Finally, graph models often fail to balance stability and plasticity, with parameter updates leading to overfitting \cite{dong2021few, kim2019edge} or weight stagnation due to insufficient updates for few-shot tasks and excessive regularization \cite{kirkpatrick2017overcoming}.

\begin{figure}[htbp]
  \centering
  \includegraphics[width=0.45\textwidth]{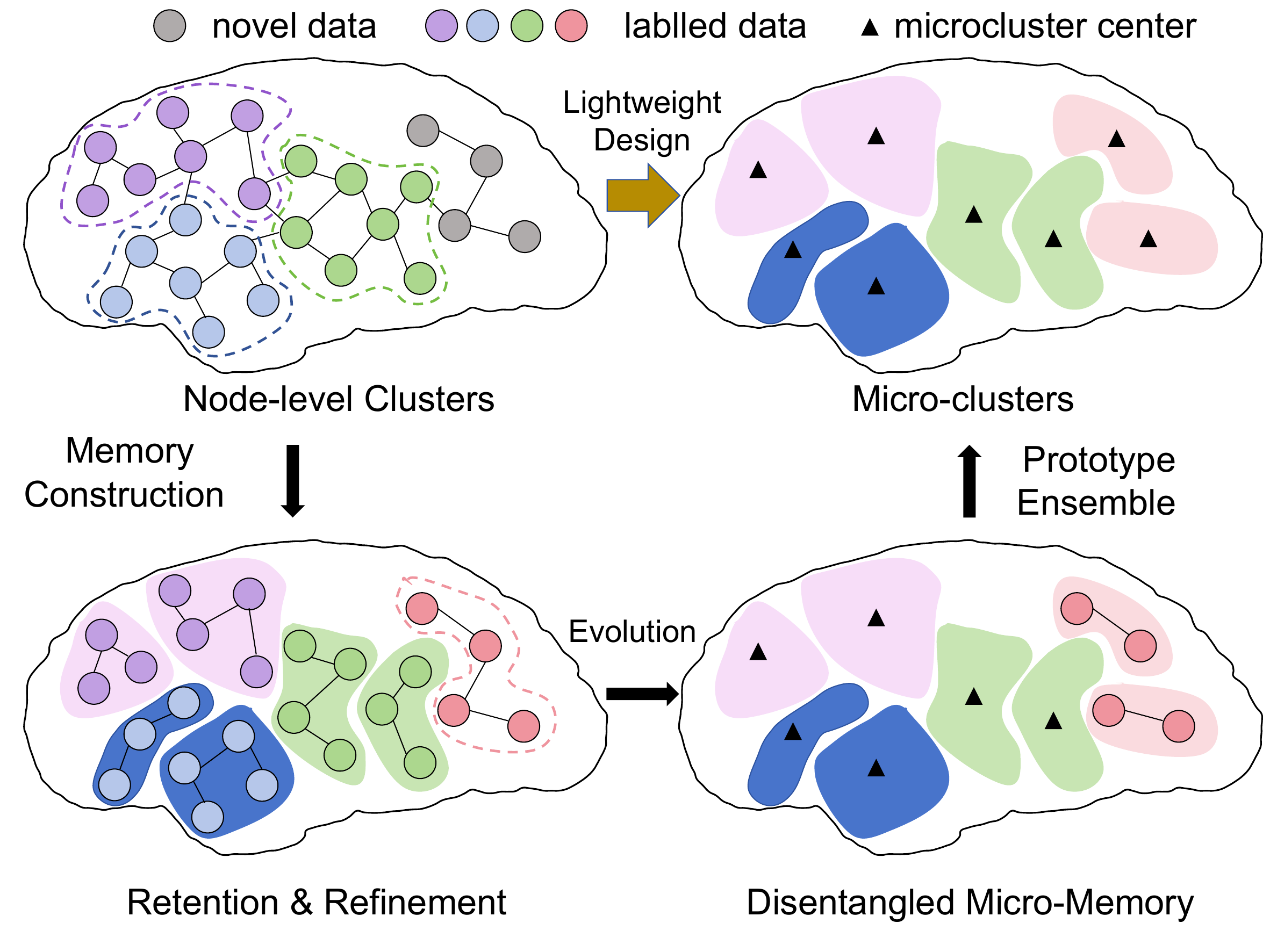}
  \caption{Hierarchical Memory Structure. Traditional methods rely on node-level data to retain knowledge, leading to inefficiency. Our method introduces a hierarchical memory structure with three stages:
(1) Retention \& Refinement: Learned nodes are clustered into micro-clusters to consolidate class representations.
(2) Disentangled Micro-Memory: New nodes are grouped into semantic-specific micro-clusters, enabling memory evolution.
(3) Lightweight Memory: Only statistical information is retained, achieving a compact and efficient memory design.}
  \label{fig:intro}
\end{figure} 

GFSCIL confronts three interconnected challenges. Catastrophic forgetting, inherent to incremental learning, is amplified in graphs due to their relational complexity, where small structural shifts can disrupt learned dependencies. Simultaneously, the plasticity-stability trade-off demands careful equilibrium: adapting to new classes without overwriting prior knowledge. Finally, label scarcity—a hallmark of few-shot learning—imposes severe constraints on feature generalizability, requiring models to infer robust node representations from minimal annotated examples. Interestingly, these issues resonate with longstanding problems in data stream clustering, where algorithms must dynamically update clusters under evolving data distributions with minimal supervision, which means both fields face similar challenges\cite{silva2013data,zubarouglu2021data}. 

Inspired by these parallels, we introduce Lightweight Plastic-Memory with Micro-Clustering framework, termed LPMC, to tackle these challenges. At the heart of our framework lies a plastic-memory module that evolves dynamically over time, continuously updating and expanding its memory to incorporate new classes while preserving previously acquired knowledge, the concept of our method is shown in Figure \ref{fig:intro}. Unlike existing methods, our memory module employs an innovative evolving micro-clustering structure, which enables the dynamic representation of class prototypes, sub-prototypes, and their interaction weights in real time. Specifically, microclusters are formed by grouping nodes around multiple cluster centers that represent sub-prototypes, while these cluster centers are further organized around a class center, representing the class prototype. This hierarchical structure ensures efficient and adaptive memory management, making our framework both lightweight and effective for dynamic graph environments.

Building on this memory module, we draw inspiration from the Model-Agnostic Meta-Learning (MAML) framework \cite{finn2017model} to propose a memory-driven meta-learning framework. Specifically, our approach integrates meta-learning in the inner loop to enhance adaptability to new tasks, while employing Graph Pseudo Incremental Learning in the outer loop to preserve stability for earlier tasks. This dual-loop design optimizes the model's ability to adapt to new tasks while maintaining a balance between stability, plasticity, and training efficiency. 

LPMC’s lightweight design is based on replacing raw data with prototypes, global structural updates with local adjustments, and retraining with meta-optimization. This is achieved through three key mechanisms. First, the Hierarchical Micro-Clustering Representation reduces memory overhead by maintaining a small set of dynamically adjusted prototypes and sub-prototypes, rather than complete samples or graph structures. Second, the Dynamic Evolution of Micro-Clusters enables the seamless integration of new knowledge by locally adjusting sub-prototypes and class prototypes during incremental phases, eliminating the need for global retraining or full graph reconstruction. Finally, the Dual-Loop Optimization mechanism enhances adaptability and stability: the inner loop uses meta-learning for rapid task adaptation, while the outer loop employs pseudo-incremental learning to stabilize old tasks through lightweight memory replay, avoiding redundant computations. Together, these mechanisms ensure an efficient and flexible framework for incremental learning.

In summary, our contributions can be outlined as follows:
\begin{itemize}
\item We propose a lightweight plastic-memory framework featuring evolving micro-clustering that dynamically organizes class prototypes and sub-prototypes through hierarchical clustering and maintains inter-class discrimination through adaptive interaction weights.

\item We design a memory-driven dual-loop framework where the inner loop implements task-specific fast adaptation via gradient meta-updates in meta-learning, while the outer loop employs graph pseudo incremental learning to consolidate structural knowledge.

\item Our proposed LPMC achieves new state-of-the-art performance and faster runtime on four major benchmark datasets under various GFSCIL scenarios.
\end{itemize}

\section{Related Work}

In recent years, several methods specifically designed to address the challenges of GFSCIL have been proposed. \cite{tan2022graph} introduces a hierarchical attention framework to balance forgetting and accuracy, but struggles with class imbalances and limited generalization. Another method leverages memory-enhanced knowledge distillation\cite{li2024efficient}, showing improved performance but facing challenges with multiple training rounds and ineffective prototype updating when labeled data is scarce.

Although existing research specifically targeting GFSCIL is still limited, valuable insights into some of its key challenges have been explored in the fields of few-shot learning and incremental learning. In graph few-shot learning, approaches can be broadly categorized into three groups: meta-learning-based, pre-training-based, and mixed methods \cite{yu2024few}. Meta-learning-based methods have been particularly influential, enabling models to adapt quickly to new tasks with limited data. These methods enhance the model's ability to capture graph structural information through node-level, edge-level, and subgraph-level adaptations, while also improving rapid adaptation capabilities through graph-level and task-level optimizations. Notable examples include: Meta-GNN \cite{zhou2019meta}, which integrates meta-learning with graph neural networks (GNNs) to create a generalizable framework independent of specific GNN architectures. G-Meta \cite{huang2020graph}, which represents nodes using local subgraphs and employs subgraph-based meta-learning. TENT \cite{wang2022task}, which adapts to new data distributions by minimizing the entropy of test-time predictions. TEG \cite{kim2023task}, which focuses on learning task-specific node embeddings. GPN \cite{ding2020graph}, which learns class prototype representations for rapid adaptation in GFSL scenarios. In contrast, pre-training-based methods leverage large-scale pre-trained models to achieve faster convergence and adaptation to specific tasks. For instance, GPPT \cite{sun2022gppt} accelerates adaptation by transforming downstream tasks into a format similar to the pre-training task using graph prompt functions. Studies have demonstrated that combining pre-trained knowledge with parameter fine-tuning yields strong performance on benchmark datasets \cite{yu2024few}.

Incremental learning is generally divided into instance, domain, and class incremental learning (CIL) \cite{luo2020appraisal}, with CIL focusing on learning new categories over time while retaining knowledge of previous ones \cite{belouadah2021comprehensive,masana2022class,zhou2024class,mittal2021essentials}. Existing CIL methods can be broadly classified into three categories: model expansion, fixed representation, and fine-tuning \cite{belouadah2021comprehensive}. Model expansion increases capacity to accommodate new knowledge; fixed representation preserves the backbone while updating the classifier; and fine-tuning modifies only the final layers. These methods include dynamic networks, which expand the model structure to adapt to data stream changes; data and parameter regularization, such as Topology-aware Weight Preserving (TWP) \cite{liu2021overcoming} and Elastic Weight Consolidation (EWC) \cite{kirkpatrick2017overcoming}, which resist forgetting by regularizing parameters or data representations; knowledge distillation methods like Learning without Forgetting (LwF) \cite{li2017learning}, which minimize the discrepancy between old and new model outputs to retain previous knowledge; data replay techniques, including Gradient Episodic Memory (GEM) \cite{lopez2017gradient} and Experience Replay GNN (ER-GNN) \cite{zhou2021overcoming}, which store previous instances and adjust learning to prevent forgetting; and model correction methods, which reduce bias in the predictions of incremental learners. This categorization illustrates the diverse strategies aimed at addressing catastrophic forgetting in CIL, each focusing on different aspects of model adaptation. Similar challenges also exist in the field of datastream clustering, where existing methods excel in real-time adaptation to evolving data distributions \cite{zubarouglu2021data,silva2013data}. For instance, methods like Chameleon \cite{xu2017dynamic} and DenStream \cite{cao2006density} employ adaptive mechanisms to handle concept drift and irregular cluster shapes, while others, such as StreamSW \cite{reddy2019streamsw}, SNCStream+ \cite{barddal2016sncstream+}, and MC-NN \cite{zhao2008real}, balance historical and recent data or enhance noise resilience. While robust and scalable, they often assume fully observable or static data, limiting their applicability to graph-structured few-shot learning.

\section{Problem Statement}

Let \(\mathcal{G} = (\mathcal{V}, \mathcal{E}, \mathbf{X})\) represent a graph, where \(\mathcal{V}\) denotes the set of nodes, \(\mathcal{E}\) denotes the set of edges, and \(\mathbf{X} \in \mathbb{R}^{|\mathcal{V}| \times d}\) represents the node feature matrix. Alternatively, the graph can be expressed as \(\mathcal{G} = \{\mathbf{A}, \mathbf{X}\}\), where \(\mathbf{A}\) is the adjacency matrix capturing the connections between nodes. In the context of \textbf{class incremental learning}, we consider a progressive sequence of learning sessions \(\mathcal{S} = \{S_0, S_1, \ldots, S_T\}\) with corresponding datasets \(\{\mathcal{D}^0, \mathcal{D}^1, \ldots, \mathcal{D}^T\}\), where \(\mathcal{D}^i = \{\mathbf{A}_{C^i}, \mathbf{X}_{C^i}\}\). Here, \(C^i\) represents the label space for session \(i\), and the label spaces are disjoint across sessions, i.e., \(C^i \cap C^j = \emptyset\) for \(i \neq j\).

The \textbf{Few-shot Class-incremental Learning (FSCIL)} scenario is defined as follows: For an \(N\)-way \(K\)-shot incremental node classification task, the first session \(S_0\) uses \(\mathcal{D}^0\) as the base dataset, providing sufficient data for conventional semi-supervised or supervised node classification training. Subsequent sessions \(S^i\) (\(i \geq 1\)) involve datasets \(\mathcal{D}^i\) (\(i \geq 1\)), which contain few-shot datasets with \(N\) novel classes, each represented by \(K\) labeled nodes. The objective is to design a model capable of maintaining strong classification performance across both base and novel classes while adapting to the evolving label space through successive learning sessions.

\begin{figure*}[tb]
    \centering 
    \includegraphics[width=1.0\textwidth]{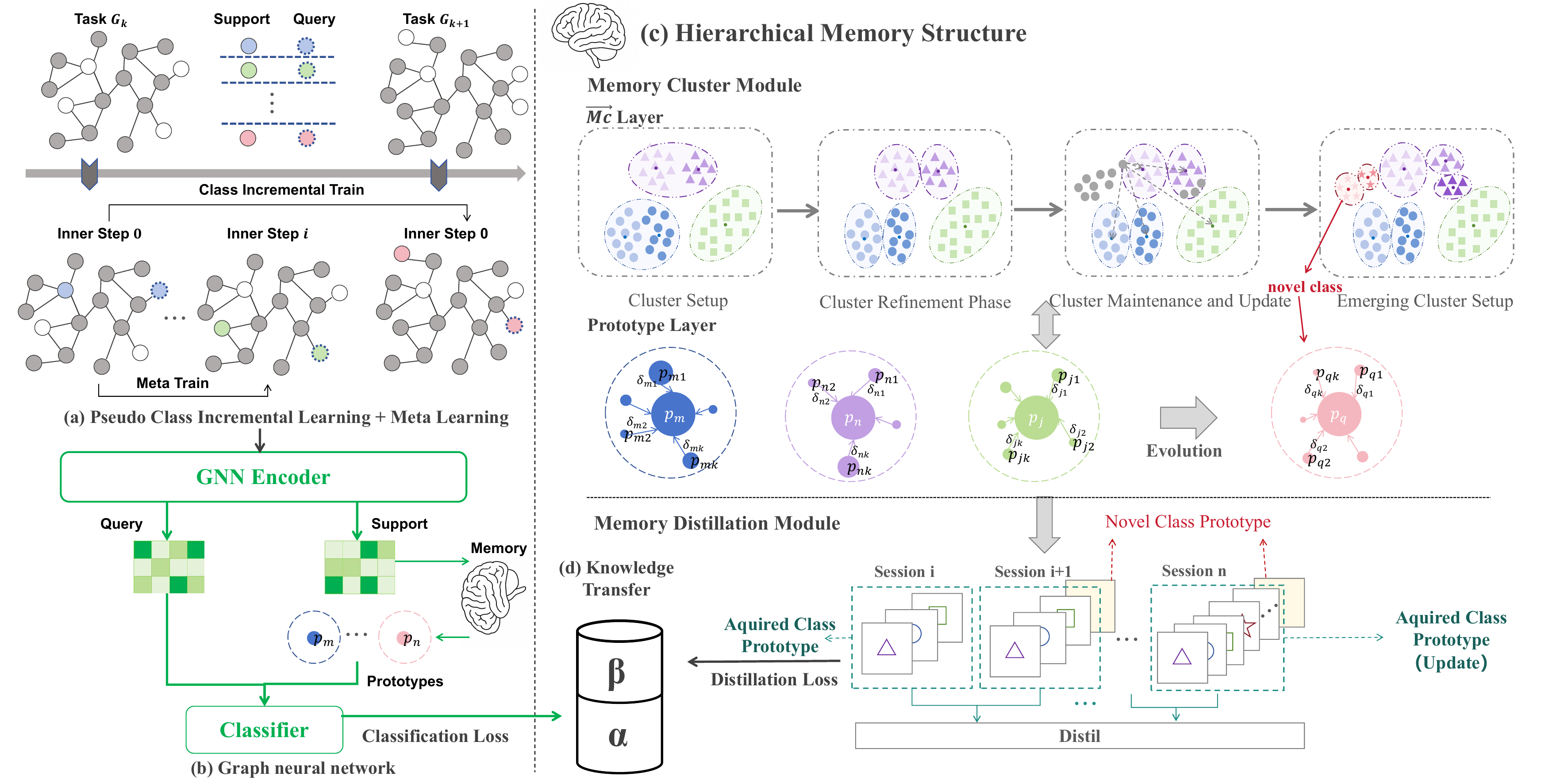}
    \caption{Overview of the LPMC framework for GFSCIL.  (a)Pseudo Class Incremental Learning with Meta-Learning: Tasks sample base and N-way pseudo novel classes. Base classes remain fixed during inner-loop meta-training, while pseudo novel classes are integrated into the base set after each session.(b)Graph neural network: Comprises a GNN encoder and a prototypical network classifier with multi-sub-prototypes.  (c)Inside Hierarchical Memory Structure, Memory Cluster Module: Constructs Mico-Clustering layer through DBSCAN and distance metric, Prototype Layer is constructed through MC layer.  Memory Distillation Module: Interact with the GNN encoder to reduce knowledge forgetting by reducing the variation of class prototypes.  (d)Knowledge Transfer: the total loss
$\mathcal{L}_{\text{total}}=\alpha\mathcal{L}_{\text{cls}}+\beta\mathcal{L}_{\text{distil}}$
(with learnable coefficients $\alpha,\beta$) is back-propagated to the GNN encoder.}
    \label{fig:framework}
\end{figure*}
\section{Methodology}
\label{sec:method}
\subsection{Pre-training Framework}

In the pre-training phase, we adopt SimGRACE \cite{xia2022simgrace}, a self-supervised contrastive learning method that leverages graph perturbation to learn structural and node-level representations. SimGRACE supports general GNN backbones such as GAT \cite{velivckovic2017graph}, GCN \cite{kipf2016semi}, and GraphSAGE \cite{zhang2019graph}. We use a 2-layer GAT as the feature extractor, which applies multi-head self-attention to aggregate neighborhood features. The propagation rules are defined as follows:
\begin{equation}
h_i^{\prime}=\sigma\left(\frac{1}{K}\sum_{k=1}^K\sum_{v_j\in\mathcal{N}(v_i)}\alpha_{ij}^kW^kh_j\right)
\end{equation}

$h_i^{\prime}$ the updated node representation,  $\sigma$ is the activation function,  $\alpha_{ij}^k$ represents the attention coefficient between nodes $v_i$ and $v_j$ for head $k,W^k$ is the weight matrix, and $K$ is the number of attention heads, where each head considers the neighbors $v_j$ of node $v_i$ in
the graph. 

The primary goal of applying contrative learning is to strengthen the alignment between augmented views of the same graph while reducing the similarity between different graphs. This process is guided by a contrastive loss function, defined as:
\begin{equation}
\mathcal{L}_{\text{contrastive}} = -\sum_{i=1}^{N} \log \frac{\exp(\text{sim}(\mathbf{z}_i, \mathbf{z}_i^+)/t)}{\sum_{j=1}^{N} \exp(\text{sim}(\mathbf{z}_i, \mathbf{z}_j)/t)}
\end{equation}

Here, z$_i$ represents the original graph-level embedding, and $\mathbf{z}_i^+$ is its Gaussian-perturbed counterpart, serving as the positive sample. The embeddings z$_j$ correspond to other nodes in the batch, acting as negative samples. By constructing positive and negative sample pairs, the model's robustness and generalization capabilities are significantly improved. The temperature parameter $t$ adjusts the model's sensitivity to differences between positive and negative pairs, while the similarity metric captures the closeness between embeddings, enabling the model to learn critical features during training.

\subsection{Plasitic-Memory Construction and Memory-Driven Training Framework}
As discussed, GFSCIL emphasizes incremental learning and mitigating forgetting. While pre-training offers basic decision-making ability, it falls short of addressing GFSCIL’s core challenges. To this end, we propose the LPMC framework (Figure~\ref{fig:framework}), which integrates a plastic-memory module and a memory-driven training scheme to improve model plasticity and stability. 

\subsubsection{Plastic-Memory Construction with Micro-Clustering} The memory module $\mathcal{M}$ consists of a micro-cluster layer and a prototype layer:
\begin{equation}
\mathcal{M} = (\mathcal{M}.mc, \mathcal{M}.prototypes)
\end{equation}
The micro-cluster layer $\mathcal{M}.mc$ ensures efficient and stable prototype updates by summarizing new and old data using statistical representations rather than storing raw node features. It models each micro-cluster as a local embedding distribution within a class, with at least one cluster per class. The prototype layer $\mathcal{M}.prototypes$ captures class-level feature representations. Implementation details follow. 

To track embedding distribution shifts, let $\{\mathcal{C}_k^{(t)}\}_{k=1}^K $ be the micro-cluster set associated with one class at time $t$, which has $K$ clusters, $K\in\mathbb{Z}^+$. For a single micro-cluster,

\begin{equation}
\mathcal{C}_k \triangleq \left(\mathbf{c}_{k},\mathbf{S}_1^k,\mathbf{S}_2^k,n_{k},r_{k}\right) \in \mathbb{R}^d \times \mathbb{R}^d \times \mathbb{R}^d \times \mathbb{N} \times \mathbb{R}^+
\end{equation}

These attributes represent the centroid of the micro-cluster, the linear sum and square sum of node embeddings, the member count, and the radius of the cluster.

These attributes are the centroid
$\mathbf{c}_k = \frac{1}{n_k}\mathbf{S}_1^k$, where $\mathbf{S}_1^k = \sum_{i=1}^{n_k} \hat{\mathbf{x}}_i$,the squared sum
$\mathbf{S}_2^k = \sum_{i=1}^{n_k}\hat{\mathbf{x}}_i^{\odot 2}$, and the adaptive radius
\begin{equation}
    r_k=\lambda\cdot\frac{1}{d}\sum_{j=1}^d\sqrt{\max\left(\frac{(\mathbf{S}_2^k)_j}{n_k}-\left(\frac{(\mathbf{S}_1^k)_j}{n_k}\right)^2,\epsilon\right)}
\end{equation}
where $d$ is the embedding dimension, $\epsilon>0$ is the preset minimum radius to ensure numerical stability, and $\lambda$ is a learnable scalar controlling the overall radius scale.

In the methodology for updating micro-clusters when new data arrives, the labeled data is determined whether belongs to an existing micro-cluster by measuring its Euclidean distance to the center of the micro-cluster relative to the radius of the micro-cluster.

For incoming sample embeddings $\mathcal{X}^{(t)}=\{\mathbf{x}_i\}_{i=1}^n$ at time $t$,  we assign them to existing micro-clusters using an adaptive radius criterion. The detailed update rules for assignment and statistical maintenance are summarized in Table~\ref{tab:micro_update}.

\begin{table}[h]
\centering
\caption{Micro-cluster update rules at time $t$}
\label{tab:micro_update}
\renewcommand{\arraystretch}{1.0}
\setlength{\tabcolsep}{6pt}
\small
\begin{tabular}{ll}
\hline
\textbf{Step} & \textbf{Update Rule} \\
\hline
Assignment & $\mathcal{A}_k^{(t)} = \left\{\mathbf{x}_i \in \mathcal{X}^{(t)} \mid \|\mathbf{x}_i - \mathbf{c}_k^{(t-1)}\|_2 \leq \rho r_k^{(t-1)}\right\}$ \\
Count update & $n_k^{(t)} = n_k^{(t-1)} + |\mathcal{A}_k^{(t)}|$ \\
Linear sum & $\mathbf{S}_1^{(k,t)} = \mathbf{S}_1^{(k,t-1)} + \sum_{\mathbf{x} \in \mathcal{A}_k^{(t)}} \mathbf{x}$ \\
Square sum & $\mathbf{S}_2^{(k,t)} = \mathbf{S}_2^{(k,t-1)} + \sum_{\mathbf{x} \in \mathcal{A}_k^{(t)}} \mathbf{x}^2$ \\
\hline
\end{tabular}
\end{table}

The remaining samples constitute the residual set
$\mathcal{X}_{\mathrm{res}}^{(t)} = \mathcal{X}^{(t)} \setminus \bigcup_{k=1}^{K} \mathcal{A}_k^{(t)}$.
We apply DBSCAN~\cite{schubert2017dbscan} to these residuals to generate new micro-clusters:
\begin{equation}
\mathcal{C}_{\mathrm{new}}^{(t)} = \left\{\mathcal{N}_{\varepsilon}^{(t)}(\mathbf{x}) \mid |\mathcal{N}_{\varepsilon}^{(t)}(\mathbf{x})| \geq n_{\mathrm{min}},\; \mathbf{x}\in\mathcal{X}_{\mathrm{res}}^{(t)}\right\},
\end{equation}
where $\mathcal{N}_{\varepsilon}(\mathbf{x})=\{\mathbf{x}^{\prime}\in\mathcal{X}_{\mathrm{res}}^{(t)}\mid\|\mathbf{x}^{\prime}-\mathbf{x}\|_{2}\leq\varepsilon\}$. 
Let $\mathbf{p}_k$ denote the centroid of micro-cluster $\mathcal{C}_k$ and $n_k$ its cardinality.
With $N_c$ the current total samples of class $c$, the relative density is $\delta_k = n_k / N_c$,
and the class prototype is computed as
\begin{equation}
\mathbf{P}_c = \sum_{k} \delta_k \cdot \mathbf{p}_k.
\end{equation}

\subsubsection{Memory Augmented class incremental learning}

We propose a dual-loop meta-learning framework comprising an outer loop with graph-based pseudo-class incremental learning (GPIL) and an inner loop for meta-training via new class simulation.

\textbf{Outer Loop: GPIL with Prototype Distillation.} The outer step implements Graph Pseudo Incremental Learning (GPIL). In GPIL, as incremental sessions proceed, the number of base classes grows while new classes shrink. To counter forgetting, we adopt a \textbf{Memory Distillation Module} with the following loss: 
\begin{equation}
    \mathcal{L}_{\mathrm{distill}}=1-\frac{1}{C}\sum_{i=1}^{C}\frac{\hat{\mathbf{p}}_{\mathrm{prev}}^{i}\cdot\hat{\mathbf{p}}_{\mathrm{curr}}^{i}}{T_\text{distil}}
\end{equation}
where $C$ is the number of selected base classes, $\hat{\mathbf{p}}_{\mathrm{prev}}^{i}$, $\hat{\mathbf{p}}_{\mathrm{curr}}^{i}$ are the normalized prototypes for the $i$-th class in the previous and current models, respectively, and $T_\text{distil}$ is the temperature parameter.

\textbf{Inner Step: Prototypical Network with Meta-Learning.} After each pseudo-incremental step, samples from the remaining new class labels are used for meta-updates. This reinforces base class knowledge and improves generalization. Model parameters are updated as
\begin{equation}
\theta = \theta - \eta\,\nabla_\theta\mathcal{L}_{\mathrm{cl}}(\theta)
         - \gamma\,\eta\,\nabla_\theta\mathcal{L}_{\mathrm{distil}}(\theta),
\end{equation}
where $\eta$ is the outer-loop learning rate and
$\nabla_\theta\mathcal{L}_{\mathrm{cl}}$, $\nabla_\theta\mathcal{L}_{\mathrm{distil}}$
are the gradients of the classification and distillation losses, respectively.

Given a query embedding $f_\theta(\mathbf{x}_i)\in\mathbb{R}^d$ and the $k$-th sub-prototype $\mathbf{p}^{(c)}_k\in\mathbb{R}^d$ of class $c$, their distance is defined as
\begin{equation}
d\!\bigl(f_\theta(\mathbf{x}_i),\mathbf{p}^{(c)}_k\bigr)=
1-\frac{f_\theta(\mathbf{x}_i)\cdot\mathbf{p}^{(c)}_k}
{\lVert f_\theta(\mathbf{x}_i)\rVert\,\lVert\mathbf{p}^{(c)}_k\rVert\,T_{\text{cl}}},
\end{equation}
where $T_{\text{cl}}$ is a temperature parameter. Each class $c$ has $K_c$ sub-prototypes with associated density weights $\{\rho^{(c)}_k\}_{k=1}^{K_c}$. For a batch of $N$ query samples $\{(\mathbf{x}_i,y_i)\}_{i=1}^{N}$ with ground-truth labels $y_i$, the classification loss is
\begin{equation}
\mathcal{L}_{\text{cl}}=-\frac{1}{N}\sum_{i=1}^{N}
\log\frac{
\exp\!\Bigl(-\sum_{k=1}^{K_{y_i}}\rho^{(y_i)}_k\,
d\!\bigl(f_\theta(\mathbf{x}_i),\mathbf{p}^{(y_i)}_k\bigr)\Bigr)
}{
\sum_{c=1}^{C}
\exp\!\Bigl(-\sum_{k=1}^{K_c}\rho^{(c)}_k\,
d\!\bigl(f_\theta(\mathbf{x}_i),\mathbf{p}^{(c)}_k\bigr)\Bigr)
},
\end{equation}
where $C$ is the total number of classes seen during training.
\begin{table}[h] 
  \caption{Statistics of evaluation datasets.}
  \label{table:datastats}
  \centering
  \small
  \begin{tabular}{c|ccccc}
  \toprule
  Dataset            & \# Nodes & \# Edges & \# Features  & Class Split    \\ \hline
  Amazon Clothing           & 24,919   & 91,680   & 9,034               & 17/30/27                       \\ 
  CoraFull              & 19,793   & 126,842   & 8,710                & 30/20/20                       \\ 
  CS &  18,333   &  163,788   & 6805               & 0/5/10                        \\ 
  Computers    &  13,752   &  491,722   & 767                & 0/5/5                        \\ 
  \bottomrule
  \end{tabular}
\end{table}
\begin{table*}[]
\caption{\label{tab: selected results}Main experiment results on the Amazon clothing, CoraFull, CS and Computers datasets under different N-way K-shot settings. Detailed results are provided in supplementary materials.}
\centering
\small
\begin{tabular}{c|ccccc|ccccc}
\hline
& \multicolumn{5}{c|}{\textbf{Amazon Clothing dataset (3-way 5-shot)}} &  \multicolumn{5}{c}{\textbf{CoraFull dataset (2-way 5-shot)}} \\ \hline
\multirow{2}{*}{Method} & \multicolumn{3}{c}{Acc. in sessions (\%) $\uparrow$} & \multirow{2}{*}{PD$\downarrow$} & \multirow{2}{*}{\shortstack{Average \\ ACC $\uparrow$}} & \multicolumn{3}{c}{Acc. in sessions (\%) $\uparrow$} & \multirow{2}{*}{PD$\downarrow$} & \multirow{2}{*}{\shortstack{Average \\ ACC $\uparrow$}} \\ \cline{2-4} \cline{7-9}
& \textbf{0} & \textbf{5} & \textbf{9} 
&  &  
& \textbf{0} & \textbf{5} & \textbf{10}  \\ 
                        \hline
                        ERGNN    & 62.40 & 29.15 & 29.48 & 32.92 & 34.87 & 73.43 & 21.84 & 11.30 & 62.13 & 29.55\\
                        GEM      & 77.54 & 27.88 & 28.82 & 48.72 & 37.11 & 69.06 & 14.78 & 7.52  & 61.54 & 22.65\\
                        MAS      & 68.70 & 27.74 & 28.91 & 39.79 & 35.17 & 69.06 & 47.04 & 46.39 & 22.67 & 49.99\\
                        LWF      & 51.97 & 18.00 & 28.75 & 23.22 & 27.80 & 73.60 & 13.91 & 7.46  & 66.14 & 24.02\\
                        EWC      & 78.26 & 29.93 & 31.92 & 46.34 & 39.84 & 69.06 & 18.59 & 12.55 & 56.51 & 27.14\\
                        TWP      & 65.83 & 25.85 & 26.86 & 38.97 & 32.48 & 69.06 & 23.29 & 13.77 & 55.29 & 27.84\\
                        Geometer & 76.28 & 30.31 & 19.91 & 56.37 & 36.82 & 72.23 & 32.79 & 16.32 & 55.91 & 35.52\\
                        HAG-Meta & \textbf{84.15} & 61.42 & 50.79 & 33.36 & 63.79 & \textbf{87.62} & 63.38 & 51.47 & 36.15 & \underline{66.57}\\
                        Mecoin   & 77.78 & 64.60 & 56.18 & \underline{21.60} & \underline{66.60} & 75.53 & 64.97 & 60.10 & \underline{15.43} & 66.22\\
                        \textbf{Ours} & 81.20 & \textbf{70.92} & \textbf{69.09} & \textbf{12.11} & \textbf{73.86} & 72.12 & \textbf{67.59} & \textbf{65.00} & \textbf{7.12} & \textbf{66.79}\\ \hline
\midrule
& \multicolumn{5}{c|}{\textbf{CS dataset (1-way 5-shot)}} &  \multicolumn{5}{c}{\textbf{Computers dataset (1-way 5-shot)}} \\ \hline
\multirow{2}{*}{Method} & \multicolumn{3}{c}{Acc. in sessions (\%) $\uparrow$} & \multirow{2}{*}{PD$\downarrow$} & \multirow{2}{*}{\shortstack{Average \\ ACC $\uparrow$}} & \multicolumn{3}{c}{Acc. in sessions (\%) $\uparrow$} & \multirow{2}{*}{PD$\downarrow$} & \multirow{2}{*}{\shortstack{Average \\ ACC $\uparrow$}} \\ \cline{2-4} \cline{7-9}
& \textbf{0} & \textbf{5} & \textbf{10} 
&  &  
& \textbf{0} & \textbf{2} & \textbf{5}  \\ 
                        \hline
ERGNN     & 100.00 & 36.93 & 30.12 & 69.88 & 38.41 & 100.00 & 33.33 & 16.67 & 83.33 & 40.83\\
GEM       & 100.00 & 17.03 & 18.09 & 81.91 & 30.30 & 100.00 & 33.33 & 16.67 & 83.33 & 40.83\\
MAS       & 100.00 & 59.54 & 63.92 & 36.08 & 60.68 & 100.00 & 33.57 & 21.64 & 78.36 & 44.75\\
LWF       & 100.00 & 36.40 & 32.51 & 67.49 & 38.30 & 100.00 & 33.33 & 16.84 & 83.16 & 40.86\\
EWC       & 100.00 & 36.62 & 38.63 & 61.37 & 38.74 & 100.00 & 33.33 & 16.67 & 83.33 & 40.83\\
TWP       & 100.00 & 47.14 & 52.02 & 47.98 & 44.44 & 100.00 & 33.33 & 16.67 & 83.33 & 40.83\\
Geometer  &  60.60 & 28.86 & 29.63 & 30.97 & 28.11 &  59.40 & 23.57 & 13.20 & 46.20 & 27.19\\
HAG-Meta  &  20.00 & 10.00 & 6.66  & \textbf{13.34} & 11.24 &  20.00 & 14.28 & 10.00 & \textbf{10.00} & 14.09\\
Mecoin    &  97.83 & 77.88 & 62.21 & 35.62 & \underline{77.50} &  91.44 & 54.94 & 67.66 & 23.78 & \underline{74.64}\\
\textbf{Ours} & \textbf{98.01} & \textbf{81.95} & \textbf{73.33} & \underline{24.68} & \textbf{84.06} & \textbf{93.60} & \textbf{91.43} & \textbf{81.00} & \underline{12.60} & \textbf{88.53}\\ \hline
\end{tabular}
\end{table*}

\section{Experiments}
\label{sec:experiments}
\subsection{Experimental Setup}

\textbf{Datasets.} Our evaluation utilizes four widely-used real-world datasets: Amazon Clothing, CoraFull, CoauthorCS, and Computers. Table \ref{table:datastats} provides the statistics and partitions of the datasets. Class split refers to the division of dataset categories based on our training framework into base classes, novel train classes, and novel test classes . Both base classes and novel train classes are accessible during the training phase, while novel test classes are only introduced during the testing phase.

\textbf{Baselines.} In evaluating our methodology, we benchmark against nine significant models to comprehensively demonstrate the effectiveness of our approach. These include three state-of-the-art methods specifically designed for GFSCIL: HAG-Meta \cite{tan2022graph}, Geometer \cite{lu2022geometer}, and Mecoin \cite{li2024efficient}. Additionally, we compare against six foundational learning frameworks tailored for graph class-incremental learning scenarios: Elastic Weight Consolidation (EWC) \cite{kirkpatrick2017overcoming}, Learning without Forgetting (LwF) \cite{li2017learning}, Topology-aware Weight Preserving (TWP) \cite{liu2021overcoming}, Gradient Episodic Memory (GEM) \cite{lopez2017gradient}, Memory Aware Synapses (MAS) \cite{aljundi2018memory}, and Experience Replay GNN (ER-GNN) \cite{zhou2021overcoming}. These comparisons aim to highlight our model's advancements in mitigating knowledge forgetting, improving accuracy, and enhancing generalization in the GFCIL setting.

\begin{table}[htbp]

\centering
\caption{Training Epochs and Running Time of Our Method and SOTA (Mecoin)}
\label{tab:method_comparison}
\small
\begin{tabular}{c|c|c|c|c|c}
\hline
Dataset & Method & Epochs & PD & Average Acc & Time \\
\hline
\multirow{2}{*}{\shortstack{Amazon \\ clothing}} & Mecoin & 2000 & 21.60 & 66.60 & 2108s \\
                                  & Ours   & 10   & 11.01 & 70.02 & 78.3s \\
\hline
\multirow{2}{*}{CoraFull}       & Mecoin & 2000 & 15.43 & 66.22 & OOM \\
                                  & Ours   & 10   & 10.57 & 65.48 & 105.8s \\
\hline
\multirow{2}{*}{computers}       & Mecoin & 2000 & 23.78 & 74.64 & 1152s \\
                                  & Ours   & 10   & 13.59 & 85.73 & 7.48s \\
\hline
\multirow{2}{*}{CS}              & Mecoin & 2000 & 35.62 & 77.50 & 1336s \\
                                  & Ours   & 10   & 21.60 & 83.22 & 12.7s \\
\hline
\end{tabular}
\end{table}

\begin{table*}[htbp]
\centering
\caption{Performance degradation (PD) and Average Accuracy under different ablation settings.}
\small
\begin{tabular}{c|c|c|cccc}
\hline
                   Pretrain & micro-cluster & meta train & Amazon\_clothing & CoraFull & computers & CS \\
\hline
\checkmark          &     \checkmark             &   \checkmark         & 12.11/73.86      & 7.12/66.79 & 12.60/88.53 & 24.68/84.06 \\
\hline
    w/o      &                  &            & (+3.68/-2.02)    & (+0.77/-1.33) & (+11.05/-12.01) & (+1.12/-0.94) \\
\hline
     &        w/o       &            & (+5.51/-1.41)    & (+3.96/-0.34) & (-0.92/-2.73) & (+2.99/-6.94) \\
\hline
        &                  &     w/o    & (+0.39/-10.01)   & (+0.55/-10.35) & (+7.38/-6.22) & (+5.96/-4.76) \\
\hline
\end{tabular}
\label{tab:ablation}
\end{table*}

\begin{table*}[h]
\caption{\label{tab:change_feature_extractor}Analysis of Backbone}
\centering

\setlength{\tabcolsep}{2.0pt}
\small
\begin{tabular}{cccccccccccccccc}
\hline
\multicolumn{15}{c}{\textbf{Amazon Clothing dataset (3-way 5-shot)}}                                                                                                                                                                                                                                                           \\ \hline
\multirow{2}{*}{Method} & \multicolumn{10}{c}{Acc. in each session ($\%$) $\uparrow$}                                                                                                                          & \multirow{2}{*}{PD$\downarrow$} & \multirow{2}{*}{\shortstack{Average \\ ACC $\uparrow$}}  & \multirow{2}{*}{} & \multirow{2}{*}{}\\ \cline{2-11}
                      & \textbf{0}     & \textbf{1}     & \textbf{2}     & \textbf{3}     & \textbf{4}     & \textbf{5}     & \textbf{6}     & \textbf{7}     & \textbf{8}     & \textbf{9}     &                     &               &    &      \\ \hline
GAT &81.20 & 80.94 & 78.93 & 77.97 & 75.81 & 70.92 & 68.24 & 68.17 & 67.30 & 69.09 & 12.11 & 73.86       & &                                                          \\
GCN  & 82.23 & 81.70 & 79.82 & 76.78 & 75.65 & 71.23 & 66.76 & 68.87 & 67.16 & 65.71 & 16.52 & 73.59     & &                                                          \\
GraphSAGE& 82.48 & 80.94 & 81.43 & 75.93 & 76.45 & 73.85 & 71.32 & 70.42 & 67.97 & 66.49 & 15.99 & 74.73  & &                                                            \\
\hline  
\end{tabular}

\smallskip

\centering

\setlength{\tabcolsep}{2.0pt}
\begin{tabular}{cccccccccccccc}
\hline
\multicolumn{14}{c}{\textbf{CoraFull dataset (2-way 5-shot)}}                                                                                                                                                                                                                                                           \\ \hline
\multirow{2}{*}{Backbone} & \multicolumn{11}{c}{Acc. in each session ($\%$) $\uparrow$}                                                                                                                          & \multirow{2}{*}{PD$\downarrow$} & \multirow{2}{*}{\shortstack{Average \\ ACC $\uparrow$}} \\ \cline{2-12}
                      & \textbf{0}     & \textbf{1}     & \textbf{2}     & \textbf{3}     & \textbf{4}     & \textbf{5}     & \textbf{6}     & \textbf{7}     & \textbf{8}     & \textbf{9}   & \textbf{10}  &                     &                       \\ \hline
GAT & 72.12 & 70.20 & 70.96 & 67.59 & 66.96 & 67.59 & 63.50 & 62.58 & 65.63 & 62.58 & 65.00 & 7.12 & 66.79                                                                    \\
GCN & 72.29 & 68.20 & 72.50 & 68.89 & 67.86 & 64.66 & 66.67 & 64.35 & 63.59 & 64.24 & 64.79 & 7.50 & 67.09                                                                       \\
GraphSAGE & 72.79 & 70.00 & 70.00 & 71.48 & 67.86 & 68.28 & 64.00 & 67.42 & 65.17 & 62.88 & 64.71 & 8.08 & 67.69                                                           \\
\hline
\end{tabular}

\end{table*}

\subsection{Main Results}

The comparative results for few-shot node classification across various datasets and settings are summarized in the Table \ref{tab: selected results}. From these results, we draw several key observations:

\textbf{Superior Performance of LPMC:} The LPMC framework consistently achieves state-of-the-art performance across all four datasets, demonstrating its effectiveness in mitigating knowledge forgetting and maintaining high accuracy in GFSCIL tasks. Specifically, LPMC successfully balances Performance Drop (PD) and average accuracy compared to other baselines. 

\textbf{Consistency Across Diverse Settings:} LPMC demonstrates consistently superior performance across all four datasets, each with distinct N-way K-shot configurations. Whether handling multi-class tasks like Amazon Clothing (3-way) and CoraFull (2-way) or single-class tasks like CS and Computers (1-way), LPMC excels in both knowledge retention and task adaptation. This consistency underscores the robustness of LPMC's design, ensuring reliable performance across diverse graph-based learning scenarios. Additional experimental results under different settings provided in supplementary materials also demonstrate LPMC’s performance in more resource-constrained scenarios.

\textbf{Comparison with Other Baselines:} While some existing models, such as Mecoin and HAG-Meta, achieve higher accuracy in initial sessions on certain datasets, their high forgetting rates significantly degrade their long-term performance. In contrast, LPMC maintains low PD values and the highest average accuracy across all sessions, outperforming these models in subsequent tasks. On the CoraFull dataset, Mecoin starts with comparable accuracy but suffers from higher PD and lower average accuracy than LPMC. 

\textbf{Efficiency and Practical Implications:} Beyond superior accuracy and consistency, LPMC demonstrates remarkable efficiency compared to baseline models. We mainly compare our method with Mecoin as it is the SOTA efficient method specially designed for GFSCIL. As explicitly quantified in Table \ref{tab:method_comparison}, our framework requires fewer training rounds and running time to achieve lower forgetting rates and higher average accuracy, making it highly suitable for practical applications with limited training resources. Even with minimal training, LPMC outperforms baselines well before reaching peak performance. Theoretical analysis of time complexity is provided in supplementary materials.

\subsection{Ablation Study}
\label{sec:ablation}
We conducted ablation studies on four datasets to evaluate the impact of pre-training, meta-training, and the micro-cluster structure on performance and forgetting. As shown in Table~\ref{tab:ablation}, the micro-clustering mechanism in LPMC yields well-separated classes and compact intra-class distributions, indicating more discriminative prototype learning. On the simpler Computers dataset, strong performance is achieved even without micro-clustering, suggesting that basic prototypes suffice. However, LPMC shows clear advantages for more complex datasets, highlighting its strength in handling intricate graph structures. Visualizations of ablation study results are provided in supplementaty materials.

\subsection{Backbone Analysis}

In our study, we evaluated the impact of three different backbones—GCN, GAT, and GraphSAGE —on the performance of our model across two datasets: Amazon Clothing (3-way 5-shot) and CoraFull (2-way 5-shot).The experimental results shown in Table \ref{tab:change_feature_extractor} reveal that the performance of the training framework is not significantly influenced by the choice of backbone, as all three architectures yield similar results in terms of forgetting rate and accuracy. These findings highlight the robustness and generalizability of our training framework, which performs effectively across different backbone architectures.

\subsection{Parameters Analysis}
We assessed model robustness with respect to two key hyper-parameters—(1) the number of inner-update steps in meta-training and (2) DBSCAN’s minPts—on the Amazon Clothing benchmark under the 3-way 5-shot setting.
\begin{figure}[htbp]
    \centering
    \includegraphics[width=0.49\textwidth]{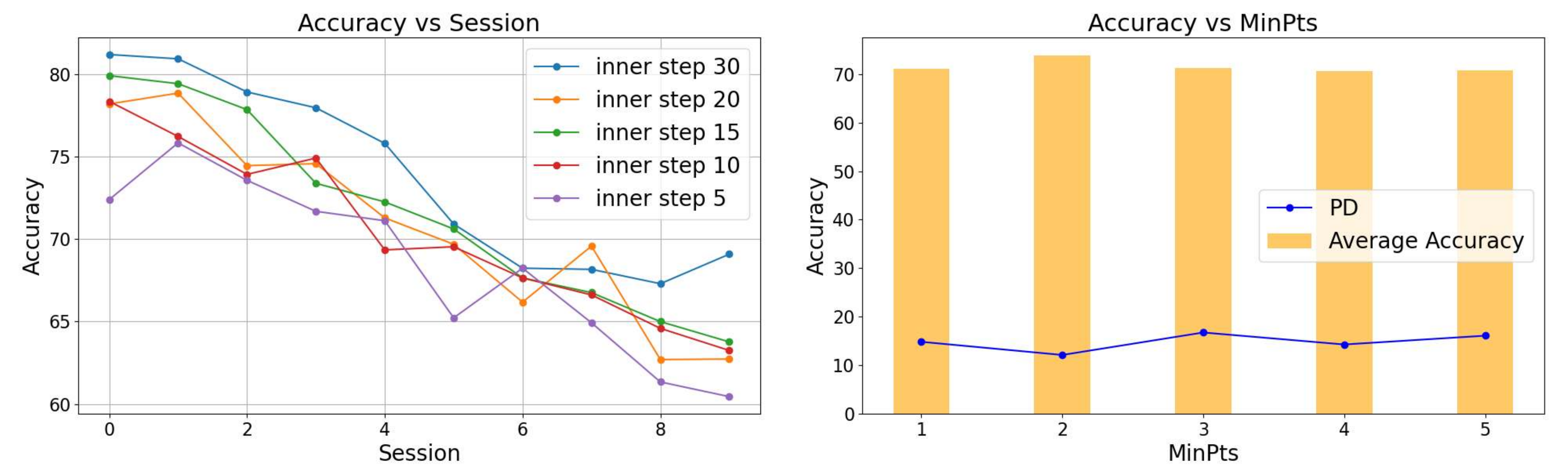}
    \caption{Impact of Inner step and MinPts on Accuracy.}
    \label{fig:para}
\end{figure} 

The inner-step count chiefly governs both initial accuracy and the final average. Adding steps, especially from very small values, yields clear gains; once the budget exceeds ~10–20 steps, however, forgetting rises slightly—likely because the outer loop is not trained long enough. Meanwhile, MinPts exerts almost no influence on either forgetting or average accuracy.

\section{Conclusion}
\label{sec:conclusion}
In this paper, we tackle the challenges of few-shot class incremental learning on dynamic graphs by proposing a lightweight plastic-memory framework.  Our novel plastic-memory module dynamically integrates new class knowledge while preserving prior knowledge, addressing the inefficiencies and high computational costs of existing methods. The micro-clustering structure enhances class node feature characterization, ensuring both stability and adaptability in evolving graph data.  Additionally, our memory-driven meta-learning framework with a dual-loop architecture improves task adaptation while maintaining performance on previously learned tasks. Extensive experiments on benchmark datasets demonstrate the framework's ability to balance stability and adaptability, with generalization error analysis confirming its robustness across different feature extractors.This work advances graph incremental learning in resource-constrained and data-scarce environments, with future research focusing on domain extension and efficiency optimization for large-scale dynamic graphs.
\newpage
\bibliography{aaai2026}

@ArtifactSoftware{R,
    title = {R: A Language and Environment for Statistical Computing},
    author = {{R Core Team}},
    organization = {R Foundation for Statistical Computing},
    address = {Vienna, Austria},
    year = {2019},
    url = {https://www.R-project.org/},
}

@inproceedings{sun2022gppt,
  title={Gppt: Graph pre-training and prompt tuning to generalize graph neural networks},
  author={Sun, Mingchen and Zhou, Kaixiong and He, Xin and Wang, Ying and Wang, Xin},
  booktitle={Proceedings of the 28th ACM SIGKDD Conference on Knowledge Discovery and Data Mining},
  pages={1717--1727},
  year={2022}
}

@article{yu2024few,
  title={Few-shot learning on graphs: from meta-learning to pre-training and prompting},
  author={Yu, Xingtong and Fang, Yuan and Liu, Zemin and Wu, Yuxia and Wen, Zhihao and Bo, Jianyuan and Zhang, Xinming and Hoi, Steven CH},
  journal={arXiv preprint arXiv:2402.01440},
  year={2024}
}

@inproceedings{zhou2019meta,
  title={Meta-gnn: On few-shot node classification in graph meta-learning},
  author={Zhou, Fan and Cao, Chengtai and Zhang, Kunpeng and Trajcevski, Goce and Zhong, Ting and Geng, Ji},
  booktitle={Proceedings of the 28th ACM International Conference on Information and Knowledge Management},
  pages={2357--2360},
  year={2019}
}

@inproceedings{xia2022simgrace,
  title={Simgrace: A simple framework for graph contrastive learning without data augmentation},
  author={Xia, Jun and Wu, Lirong and Chen, Jintao and Hu, Bozhen and Li, Stan Z},
  booktitle={Proceedings of the ACM Web Conference 2022},
  pages={1070--1079},
  year={2022}
}

@inproceedings{ding2020graph,
  title={Graph prototypical networks for few-shot learning on attributed networks},
  author={Ding, Kaize and Wang, Jianling and Li, Jundong and Shu, Kai and Liu, Chenghao and Liu, Huan},
  booktitle={Proceedings of the 29th ACM International Conference on Information \& Knowledge Management},
  pages={295--304},
  year={2020}
}

@article{huang2020graph,
  title={Graph meta learning via local subgraphs},
  author={Huang, Kexin and Zitnik, Marinka},
  journal={Advances in neural information processing systems},
  volume={33},
  pages={5862--5874},
  year={2020}
}

@inproceedings{wang2022task,
  title={Task-adaptive few-shot node classification},
  author={Wang, Song and Ding, Kaize and Zhang, Chuxu and Chen, Chen and Li, Jundong},
  booktitle={Proceedings of the 28th ACM SIGKDD Conference on Knowledge Discovery and Data Mining},
  pages={1910--1919},
  year={2022}
}

@inproceedings{kim2023task,
  title={Task-equivariant graph few-shot learning},
  author={Kim, Sungwon and Lee, Junseok and Lee, Namkyeong and Kim, Wonjoong and Choi, Seungyoon and Park, Chanyoung},
  booktitle={Proceedings of the 29th ACM SIGKDD Conference on Knowledge Discovery and Data Mining},
  pages={1120--1131},
  year={2023}
}

@ARTICLE{9416834,
  author={Xia, Feng and Sun, Ke and Yu, Shuo and Aziz, Abdul and Wan, Liangtian and Pan, Shirui and Liu, Huan},
  journal={IEEE Transactions on Artificial Intelligence}, 
  title={Graph Learning: A Survey}, 
  year={2021},
  volume={2},
  number={2},
  pages={109-127},
  doi={10.1109/TAI.2021.3076021}}

@article{luo2020appraisal,
  title={An appraisal of incremental learning methods},
  author={Luo, Yong and Yin, Liancheng and Bai, Wenchao and Mao, Keming},
  journal={Entropy},
  volume={22},
  number={11},
  pages={1190},
  year={2020},
  publisher={MDPI}
}

@article{yuan2024survey,
  title={A survey on continual semantic segmentation: Theory, challenge, method and application},
  author={Yuan, Bo and Zhao, Danpei},
  journal={IEEE Transactions on Pattern Analysis and Machine Intelligence},
  year={2024},
  publisher={IEEE}
}

@inproceedings{finn2017model,
  title={Model-agnostic meta-learning for fast adaptation of deep networks},
  author={Finn, Chelsea and Abbeel, Pieter and Levine, Sergey},
  booktitle={International conference on machine learning},
  pages={1126--1135},
  year={2017},
  organization={PMLR}
}

@inproceedings{tan2022graph,
  title={Graph few-shot class-incremental learning},
  author={Tan, Zhen and Ding, Kaize and Guo, Ruocheng and Liu, Huan},
  booktitle={Proceedings of the fifteenth ACM international conference on web search and data mining},
  pages={987--996},
  year={2022}
}

@article{belouadah2021comprehensive,
  title={A comprehensive study of class incremental learning algorithms for visual tasks},
  author={Belouadah, Eden and Popescu, Adrian and Kanellos, Ioannis},
  journal={Neural Networks},
  volume={135},
  pages={38--54},
  year={2021},
  publisher={Elsevier}
}

@article{tian2024survey,
  title={A survey on few-shot class-incremental learning},
  author={Tian, Songsong and Li, Lusi and Li, Weijun and Ran, Hang and Ning, Xin and Tiwari, Prayag},
  journal={Neural Networks},
  volume={169},
  pages={307--324},
  year={2024},
  publisher={Elsevier}
}

@article{li2024efficient,
  title={An Efficient Memory Module for Graph Few-Shot Class-Incremental Learning},
  author={Li, Dong and Zhang, Aijia and Gao, Junqi and Qi, Biqing},
  journal={arXiv preprint arXiv:2411.06659},
  year={2024}
}

@inproceedings{cao2006density,
  title={Density-based clustering over an evolving data stream with noise},
  author={Cao, Feng and Estert, Martin and Qian, Weining and Zhou, Aoying},
  booktitle={Proceedings of the 2006 SIAM international conference on data mining},
  pages={328--339},
  year={2006},
  organization={SIAM}
}

@inproceedings{lu2022geometer,
  title={Geometer: Graph few-shot class-incremental learning via prototype representation},
  author={Lu, Bin and Gan, Xiaoying and Yang, Lina and Zhang, Weinan and Fu, Luoyi and Wang, Xinbing},
  booktitle={Proceedings of the 28th ACM SIGKDD conference on knowledge discovery and data mining},
  pages={1152--1161},
  year={2022}
}

@article{kirkpatrick2017overcoming,
  title={Overcoming catastrophic forgetting in neural networks},
  author={Kirkpatrick, James and Pascanu, Razvan and Rabinowitz, Neil and Veness, Joel and Desjardins, Guillaume and Rusu, Andrei A and Milan, Kieran and Quan, John and Ramalho, Tiago and Grabska-Barwinska, Agnieszka and others},
  journal={Proceedings of the national academy of sciences},
  volume={114},
  number={13},
  pages={3521--3526},
  year={2017},
  publisher={National Acad Sciences}
}

@article{li2017learning,
  title={Learning without forgetting},
  author={Li, Zhizhong and Hoiem, Derek},
  journal={IEEE transactions on pattern analysis and machine intelligence},
  volume={40},
  number={12},
  pages={2935--2947},
  year={2017},
  publisher={IEEE}
}

@inproceedings{liu2021overcoming,
  title={Overcoming catastrophic forgetting in graph neural networks},
  author={Liu, Huihui and Yang, Yiding and Wang, Xinchao},
  booktitle={Proceedings of the AAAI conference on artificial intelligence},
  pages={8653--8661},
  year={2021}
}

@article{lopez2017gradient,
  title={Gradient episodic memory for continual learning},
  author={Lopez-Paz, David and Ranzato, Marc'Aurelio},
  journal={Advances in neural information processing systems},
  volume={30},
  year={2017}
}

@inproceedings{aljundi2018memory,
  title={Memory aware synapses: Learning what (not) to forget},
  author={Aljundi, Rahaf and Babiloni, Francesca and Elhoseiny, Mohamed and Rohrbach, Marcus and Tuytelaars, Tinne},
  booktitle={Proceedings of the European conference on computer vision (ECCV)},
  pages={139--154},
  year={2018}
}

@inproceedings{zhou2021overcoming,
  title={Overcoming catastrophic forgetting in graph neural networks with experience replay},
  author={Zhou, Fan and Cao, Chengtai},
  booktitle={Proceedings of the AAAI Conference on Artificial Intelligence},
  pages={4714--4722},
  year={2021}
}

@article{masana2022class,
  title={Class-incremental learning: survey and performance evaluation on image classification},
  author={Masana, Marc and Liu, Xialei and Twardowski, Bart{\l}omiej and Menta, Mikel and Bagdanov, Andrew D and Van De Weijer, Joost},
  journal={IEEE Transactions on Pattern Analysis and Machine Intelligence},
  volume={45},
  number={5},
  pages={5513--5533},
  year={2022},
  publisher={IEEE}
}

@article{zhou2024class,
  title={Class-incremental learning: A survey},
  author={Zhou, Da-Wei and Wang, Qi-Wei and Qi, Zhi-Hong and Ye, Han-Jia and Zhan, De-Chuan and Liu, Ziwei},
  journal={IEEE Transactions on Pattern Analysis and Machine Intelligence},
  year={2024},
  publisher={IEEE}
}

@inproceedings{mittal2021essentials,
  title={Essentials for class incremental learning},
  author={Mittal, Sudhanshu and Galesso, Silvio and Brox, Thomas},
  booktitle={Proceedings of the IEEE/CVF Conference on Computer Vision and Pattern Recognition},
  pages={3513--3522},
  year={2021}
}

@inproceedings{kim2019edge,
  title={Edge-labeling graph neural network for few-shot learning},
  author={Kim, Jongmin and Kim, Taesup and Kim, Sungwoong and Yoo, Chang D},
  booktitle={Proceedings of the IEEE/CVF conference on computer vision and pattern recognition},
  pages={11--20},
  year={2019}
}

@article{zhou2022few,
  title={Few-shot class-incremental learning by sampling multi-phase tasks},
  author={Zhou, Da-Wei and Ye, Han-Jia and Ma, Liang and Xie, Di and Pu, Shiliang and Zhan, De-Chuan},
  journal={IEEE Transactions on Pattern Analysis and Machine Intelligence},
  volume={45},
  number={11},
  pages={12816--12831},
  year={2022},
  publisher={IEEE}
}

@inproceedings{tao2020few,
  title={Few-shot class-incremental learning},
  author={Tao, Xiaoyu and Hong, Xiaopeng and Chang, Xinyuan and Dong, Songlin and Wei, Xing and Gong, Yihong},
  booktitle={Proceedings of the IEEE/CVF conference on computer vision and pattern recognition},
  pages={12183--12192},
  year={2020}
}

@inproceedings{dong2021few,
  title={Few-shot class-incremental learning via relation knowledge distillation},
  author={Dong, Songlin and Hong, Xiaopeng and Tao, Xiaoyu and Chang, Xinyuan and Wei, Xing and Gong, Yihong},
  booktitle={Proceedings of the AAAI Conference on Artificial Intelligence},
  volume={35},
  number={2},
  pages={1255--1263},
  year={2021}
}

@inproceedings{chi2022metafscil,
  title={Metafscil: A meta-learning approach for few-shot class incremental learning},
  author={Chi, Zhixiang and Gu, Li and Liu, Huan and Wang, Yang and Yu, Yuanhao and Tang, Jin},
  booktitle={Proceedings of the IEEE/CVF conference on computer vision and pattern recognition},
  pages={14166--14175},
  year={2022}
}

@article{snell2017prototypical,
  title={Prototypical networks for few-shot learning},
  author={Snell, Jake and Swersky, Kevin and Zemel, Richard},
  journal={Advances in neural information processing systems},
  volume={30},
  year={2017}
}

@article{schubert2017dbscan,
  title={DBSCAN revisited, revisited: why and how you should (still) use DBSCAN},
  author={Schubert, Erich and Sander, J{\"o}rg and Ester, Martin and Kriegel, Hans Peter and Xu, Xiaowei},
  journal={ACM Transactions on Database Systems (TODS)},
  volume={42},
  number={3},
  pages={1--21},
  year={2017},
  publisher={Acm New York, NY, USA}
}

@article{silva2013data,
  title={Data stream clustering: A survey},
  author={Silva, Jonathan A and Faria, Elaine R and Barros, Rodrigo C and Hruschka, Eduardo R and Carvalho, Andr{\'e} CPLF de and Gama, Jo{\~a}o},
  journal={ACM Computing Surveys (CSUR)},
  volume={46},
  number={1},
  pages={1--31},
  year={2013},
  publisher={ACM New York, NY, USA}
}

@article{zubarouglu2021data,
  title={Data stream clustering: a review},
  author={Zubaro{\u{g}}lu, Alaettin and Atalay, Volkan},
  journal={Artificial Intelligence Review},
  volume={54},
  number={2},
  pages={1201--1236},
  year={2021},
  publisher={Springer}
}

@article{xu2017dynamic,
  title={Dynamic chameleon authentication tree for verifiable data streaming in 5G networks},
  author={Xu, Jian and Li, Fuxiang and Chen, Ke and Zhou, Fucai and Choi, Junho and Shin, Juhyun},
  journal={IEEE Access},
  volume={5},
  pages={26448--26459},
  year={2017},
  publisher={IEEE}
}

@article{reddy2019streamsw,
  title={StreamSW: A density-based approach for clustering data streams over sliding windows},
  author={Reddy, K Shyam Sunder and Bindu, C Shoba},
  journal={Measurement},
  volume={144},
  pages={14--19},
  year={2019},
  publisher={Elsevier}
}

@article{barddal2016sncstream+,
  title={SNCStream+: Extending a high quality true anytime data stream clustering algorithm},
  author={Barddal, Jean Paul and Gomes, Heitor Murilo and Enembreck, Fabr{\'\i}cio and Barth{\`e}s, Jean-Paul},
  journal={Information Systems},
  volume={62},
  pages={60--73},
  year={2016},
  publisher={Elsevier}
}

@article{zhao2008real,
  title={Real-time feature selection in traffic classification},
  author={Zhao, Jing-jing and Huang, Xiao-hong and Qiong, SUN and Yan, MA},
  journal={The Journal of China Universities of Posts and Telecommunications},
  volume={15},
  pages={68--72},
  year={2008},
  publisher={Elsevier}
}

@article{zhang2019graph,
  title={Graph convolutional networks: a comprehensive review},
  author={Zhang, Si and Tong, Hanghang and Xu, Jiejun and Maciejewski, Ross},
  journal={Computational Social Networks},
  volume={6},
  number={1},
  pages={1--23},
  year={2019},
  publisher={Springer}
}

@article{velivckovic2017graph,
  title={Graph attention networks},
  author={Veli{\v{c}}kovi{\'c}, Petar and Cucurull, Guillem and Casanova, Arantxa and Romero, Adriana and Lio, Pietro and Bengio, Yoshua},
  journal={arXiv preprint arXiv:1710.10903},
  year={2017}
}

@article{kipf2016semi,
  title={Semi-supervised classification with graph convolutional networks},
  author={Kipf, Thomas N and Welling, Max},
  journal={arXiv preprint arXiv:1609.02907},
  year={2016}
}
\newpage

\end{document}